\documentclass[10pt, a4paper]{article}
\usepackage{booktabs}
\usepackage{multirow}
\usepackage{subcaption}
\usepackage{float} % Required for the [H] placement
\usepackage{relsize}

\usepackage{siunitx}

\usepackage[final]{LREC2026} % this is the new style
\title{EPIC-EuroParl-UdS:\\ Information-Theoretic Perspectives on Translation and Interpreting}

\name{Maria Kunilovskaya, Christina Pollkläsener} 

\address{Saarland University, University of Hildesheim \\
         maria.kunilovskaya@uni-saarland.de, pollklaesener@uni-hildesheim.de}

\abstract{
This paper introduces an updated and combined version of the bidirectional English-German EPIC-UdS (spoken) and EuroParl-UdS (written) corpora containing original European Parliament speeches as well as their translations and interpretations.
The new version corrects metadata and text errors identified through previous use, refines the content, updates linguistic annotations, and adds new layers, including word alignment and word-level surprisal indices.
The combined resource is designed to support research using information-theoretic approaches to language variation, particularly studies comparing written and spoken modes, and examining disfluencies in speech, as well as traditional translationese studies, including parallel (source vs. target) and comparable (original vs. translated) analyses.
The paper outlines the updates introduced in this release, summarises previous results based on the corpus, and presents a new illustrative study. The study validates the integrity of the rebuilt spoken data and evaluates probabilistic measures derived from base and fine-tuned GPT-2 and machine translation models on the task of filler particles prediction in interpreting.
 \\ \newline  }

\begin{document}

\maketitleabstract

\section{Introduction}\label{sec:intro}
In recent years, information-theoretic approaches have gained traction in translation and interpreting studies as a way to explore processing effort~\cite{Carl2017, teichTranslationInformationTheory2020, Wei2022} or difficulty~\cite{lim-etal-2024-predicting} and linguistic variation, especially triggered by modality (written or spoken)~\cite{Nikolaev2020, LapshinovaEtAL2022PBML, przybyl2022_epic1, kunilovskaya2025information}. Yet, researchers in this area often lack ready-to-use resources annotated with word-level information-theoretic indices. Therefore, they either rely on corpus frequencies, i.e., unigram probabilities that ignore context, or must generate such data themselves, which can be time-consuming and resource-intensive. 

The updated EPIC-UdS and EuroParl-UdS corpora, combined into EPIC-EuroParl-UdS corpus for English$\leftrightarrow$German (DE$\leftrightarrow$EN, i.e., DE-EN, EN-DE), address this gap by providing word-level surprisal values derived from both off-the-shelf (or base) and fine-tuned neural models. The underlying GPT-2 data have been used in several studies, demonstrating the usefulness of surprisal for addressing questions in process-oriented interpreting research and in studies of linguistic variation across spoken and written modes~\cite{kunilovskaya2025information, pollklaesener2025surprisal, pollklaesener2025euh}. This paper extends the corpus by adding per-word surprisal from machine translation (MT) models and examines their performance on predicting filler particles (FPs, \textit{euh, hum, hm}) in interpreting.

The combined corpus builds on the previously released versions~\citet{karakanta2018europarl} for the written subcorpus and~\citet{PrzybylEtAl2022lrec} for the spoken subcorpus, recycling their content to varying degrees. 

Section~\ref{sec:related} places the updated corpus among existing translation and interpreting corpora and introduces key concepts and benefits of information theory. Section~\ref{ssec:structure} describes the data structure, formats and availability of the corpus. Section~\ref{ssec:sources} explains how the prior work was taken into account and how the current release differs from previous versions. Sections~\ref{ssec:prepro} and~\ref{ssec:anno} provide details on preprocessing and annotation. Section~\ref{sec:applications} presents applications of the new corpus, including a novel study using and evaluating the new annotation layers (MT surprisal, word alignment, fine-tuned surprisal) to predict FPs in interpreting. 

\section{Background}\label{sec:related}
Speeches from the European Parliament have long served as a rich data source for translation and interpreting studies. The EuroParl Corpus~\cite{koehn-2005-europarl} remains one of the most influential written parallel resources for numerous language pairs, even though it only provides scarce metadata. EuroParl-UdS~\cite{karakanta2018europarl} tries to remedy this by re-crawling the speeches and adapting the data for translationese studies, offering a filter based on the speakers' nationality as an approximation for their native language. \citet{amponsah-kaakyire-etal-2021-europarl-b6} use the same re-crawling-pipeline as~\citet{karakanta2018europarl} but align text paragraph-wise, which impedes fine-grained analysis.  
Our updated EPIC-EuroParl-UdS corpus includes document-, sentence- and word alignment. On the spoken side, the EPIC corpora family~\cite{Bernardini2018_epicfamily}, which includes EPIC-UdS~\cite{PrzybylEtAl2022lrec}, relies on manual transcriptions from audio files with European Parliament speeches. Recently, Automatic Speech Recognition (ASR) technologies have been used to accelerate transcription~\cite{fedotova-etal-2024-eptic-update}. However, disfluencies transcription still requires manual revision since disfluencies are not yet reliably detected by ASR systems. Our updated corpus is unique in its incorporation of surprisal values for both translation and interpreting data and their sources. 

Surprisal is a measure rooted in information theory~\cite{Shannon1948} that captures the (un)expectedness of a given word based on its probability in context according to a language model. It is considered to be an indicator of cognitive effort required to comprehend or produce language~\cite{Hale2001, teichTranslationInformationTheory2020}. Unexpected (high surprisal) words are associated with, e.g., longer word durations (e.g.~\citealt{malisz2018_worddurationsurprisal}), longer reading times (e.g.~\citealt{wilcox2023reading}) and more disfluencies (e.g.~\citealt{ pollklaesener2025surprisal}). 

In translation and interpreting studies, surprisal and other information-theoretic measures have been used in two types of process studies. First, in experimental settings, surprisal has been correlated with behavioural data such as reading times, keystroke logging, and predictive eye movements (e.g.~\citealp{Carl2017, hodzik2017predictive, Wei2023, lim-etal-2024-predicting}). Second, in naturalistic production studies, these measures have been used to annotate and analyse texts in parallel and comparable corpora (e.g.,~\citealt{MartinezTeich2017, Nikolaev2020, LapshinovaEtAL2022PBML, Kunilovskaya2023parallel}).

Most of these studies use n-gram surprisal, including unigram surprisal, estimated from the very corpus under analysis. It is our hope that our corpus will enable information-theoretic research using more robust modelling approaches, in which the corpus under analysis serves as held-out test data rather than the source of probability estimates.

\section{Corpus Design and Construction}
This section describes how previous EuroParl-based resources were updated and extended to create \textit{EPIC-EuroParl-UdS}. 
This update was motivated by several factors. 

First, we aimed to ensure consistent annotation and formatting across the spoken and written components to eliminate any preprocessing-related discrepancies. The previous release of EuroParl-UdS lacked a non-tokenised raw version, and EPIC-UdS appeared without punctuation or interpreter IDs.

The two components had dissimilar file- and segment-naming conventions, data formats and contained incomplete metadata.  % date formats fixed, speaker_name formatting is harmonised

Second, we excluded all English–Spanish data, making the new resource strictly bidirectional English$\leftrightarrow$German, as the Spanish data were not comparable in size or directionality across modalities.

Third, the written and spoken components shared overlapping material (speeches appearing in both forms), with the degree of overlap varying substantially across translation directions: 145 speeches from DE-EN spoken subcorpus also appear in written, while in EN-DE only five documents are shared. To prevent bias in direction-wise cross-modal comparisons, we removed overlapping documents from the larger written component. 

Fourth, the EuroParl-UdS data were imbalanced across translation directions, with DE–EN containing roughly 55\% more segment pairs.

Such an imbalance introduces bias in translationese studies, where sources in one translation direction are used as comparable originals for targets in the other direction. In the new corpus, the written component is balanced across translation directions. Additionally, we improved the automatic sentence alignment in the written subcorpus by using domain-specific glossaries and introducing a quality cut-off value to exclude poorly aligned document pairs. The resulting alignment quality was verified manually.

Finally, we considered extending the written corpus with debates translated after July 2018, the date of the latest speech included in the previous release. However, the new data could not be retrieved using the existing EuroParl-UdS extraction pipeline without compromising the continuity of the original data collection principles. Moreover, as of 1 November 2023, the European Parliament ceased to publicly release official translations as part of the verbatim reports.\footnote{https://www.europarl.europa.eu/doceo/document/RULES-9-2023-07-10-RULE-204\_EN.html}

\subsection{Structure and Availability}\label{ssec:structure}

The EPIC-EuroParl-UdS Corpus comprises two subcorpora--\textit{written and spoken}--each distributed in three complementary data formats.\footnote{Corpus: https://zenodo.org/records/19010272; DOI: 10.5281/zenodo.19010272\\ Code repository: https://github.com/SFB1102/b7-lrec2026} Each data format is a set of compressed plain-text, UTF-8 encoded tabular files (.tsv.gz), where rows correspond to textual units, and columns contain the associated text, metadata, annotations, or numeric values. Each format emphasises a different annotation layer, while item identifiers remain consistent across all tables.

\begin{description}
    \item[vertical] This word-level format is accepted as primary and is based on Stanza tokenisation (see Section~\ref{ssec:prepro}). It includes \texttt{word\_id}, 10 standard UD annotation fields, word-level surprisal values from GPT-2 and MT models, and word alignments, along with a number of key metadata useful for sorting the data. 
    
    \item[long] In this format, each row represents a segment (\texttt{seg}) and columns store the corresponding metadata, as well as segment-level GPT-2 surprisal aggregates, i.e. average segment surprisal values (gpt\_AvS). 
    
    \item[wide] This format provides a parallel, segment-level view for each subcorpus and features aggregates of MT indices (mt\_AvS).
\end{description}

Some columns may contain N/A values including predominantly string-based fields such as \textit{token, pos, lemma}, etc. in the vertical format. These arise from alignment asymmetries (e.g. omissions or additions in translation/interpreting) where no corresponding material exists in the source/target. 
N/A values are also used for words that do not receive surprisal scores, such as FPs, expanded contractions, or surprisal annotation errors.

The item ID pattern is <ttype>\_<mode>\_\allowbreak<src\_lang>\_<tgt\_lang>\_<doc\_id>-<seg\_id>:\allowbreak<word\_id>. For example, ORG\_SP\_DE\_EN\_131-02:001 refers to the first word (001) in the second segment (02) of the original (ORG) German document (131), interpreted into English within the spoken subcorpus (SP).

The \texttt{speaker\_id} attribute warrants special attention. These identifiers are particularly valuable for regression-based analyses, where speaker and/or document IDs can be incorporated as random effects to account for inter-speaker and inter-document variability.
In the written subcorpus, the \textit{speaker\_id} field corresponds to the \textit{name}; however, the identities of translators are not available (N/A is filled for all absent values). In the spoken subcorpora, the \textit{speaker\_id} field contains speaker names for original speeches, whereas interpreters are identified by newly introduced numeric codes (e.g., \textit{fDE1}) based on manual voice analysis and gender-based verification.

The spoken corpora in all formats appear in two versions: a \textit{clean} version without FPs and segment pairs where either source or target was empty and a full version (\textit{fp\_none}) including these features. 

Additional document-level metadata inherited from earlier releases (e.g. nationality, birth\_date, birth\_place, n\_party, etc.) are made available in one separate file per subcorpus.

Appendix A has a snapshot of the EPIC-EuroParl-UdS corpus file structure (Figure~\ref{fig:structure}) as well as lists of text and metadata fields associated with each format (see Table~\ref{tab:fields}). 

\subsection{Data Collection, Bitext and Filtering}\label{ssec:sources}
Initial text extraction and bitext creation are described separately for the spoken and written data, as each data source format required a number of subcorpus-specific steps.

\paragraph{Interpreting subcorpus: EPIC-UdS.}
The current corpus release introduces newly realigned and punctuated data based on the previously manually transcribed material, preserving eight types of disfluencies (see~\citealp{PrzybylEtAl2022lrec} for details). Adding punctuation improved comparability with written data and enabled more reliable downstream processing.

A fragment from a spoken transcript with phonetic annotations of disfluencies is shown below:
\begin{quote}
\small
and finally, / \textit{hum} / I'm [1\#I am] seeking to / \textit{euh} take out / the s/ [s:] the ad/ dition [2\#addition] of split and \textit{hm} separate [s:eperate] v/ ow/ votes [v:otes] [3\#] / to [to:] the procedure that will permit / the President to refer \textit{euh} / back to a [a:] \textit{euh} / committee, / a r/ f/ f/ f/ f/ f/ report / which has attracted m/ ow/ m/ more [4\#] than \textit{euh} f/ fifty [f:ifty] [2\#] substantive a/ a/ a/ am/ m/ mendments [6\#amendments].
\end{quote}
A preliminary analysis of the distribution of annotation categories indicated that some categories were inconsistently transcribed. For the sake of reliability, only FPs were preserved in the combined corpus. However, the segment-level spoken data includes the total counts of all annotated disfluencies (\texttt{disfluencies}),\footnote{Monospace script is used throughout the paper to refer to column names in the released tabular data.} counts for the three more reliably annotated categories, namely \textit{FPs, midword breaks, repetitions} and \textit{truncations} (\texttt{fillers+3}) and counts of FPs only (\texttt{fillers}, including counts of \textit{euh, hum, hm} forms).

After text preprocessing (standardising forms, mapping phonetic variants and removing all annotated disfluencies except FPs), the segment appears in the released corpus as follows:

\begin{quote}
\small
And finally, \textit{hum} I am seeking to \textit{euh} take out the addition of split and \textit{hm} separate votes to the procedure that will permit the President to refer \textit{euh} back to a \textit{euh} committee, a report which has attracted more than \textit{euh} fifty substantive amendments.
\end{quote}

Beyond disfluencies, another spoken-specific characteristic concerns segment alignment.
Unlike its written counterpart, the spoken data contains empty source or target segments, reflecting cases where interpreters omitted or added content.

Depending on the experimental setup, empty or very short segments (fewer than N words) can be treated differently. For experiments where parallelism between source and target is not required (e.g., comparing DE originals and DE targets), such segments can be filtered independently from originals and targets. For experiments requiring strict parallelism, the entire source-target pairs have to be dropped where either side is empty or too short. This will result in a dataset up to 10\% smaller than the initial spoken subcorpus.

The quantitative parameters of spoken-specific features are summarised in Table~\ref{tab:spoken_features}.

\begin{table}[h!]
\centering
\resizebox{\columnwidth}{!}{%
\begin{tabular}{@{}lllrrr@{}}
\toprule
                                        lpair & ttype  & segs & \% empty  & FP & \% segs\_w/FP \\ 
\midrule
\multirow{2}{*}{DE-EN}  & source & \multirow{2}{*}{3,249}       & 2.2  & 632      & 12.6   \\
                       & target &                               & 8.5 & 2,328     & 36.6  \\
\multirow{2}{*}{EN-DE}  & source & \multirow{2}{*}{3,440}       & 2.4  & 1,217     & 23.4   \\
                        & target &                               & 7.5 & 3,255     & 44.4  \\ 
\bottomrule
\end{tabular}
}
\caption{\label{tab:spoken_features}Ratios of empty segments and segments with filler particles (FPs), and total counts of FPs in spoken data by text type and translation direction.}
\end{table}
Segments containing FPs account for more than one-third of interpreted segments in both translation directions, with an average of approximately two FPs per segment. 

\paragraph{Translation subcorpus: EuroParl-UdS.}
The officially released written source speeches and their translations (up to July 2018) were re-extracted from XML files downloaded from the European Parliament website using the EuroParl-UdS scripts.\footnote{https://github.com/hut-b7/europarl-uds}
We also applied the existing filter to retain only original speeches delivered by native speakers of English or German, ensuring greater reliability of translationese comparisons between original and translated language.

The extracted data were re-aligned using LF Aligner~\citelanguageresource{lfaligner}
, a wrapper around the \textit{Hunalign} library~\cite{varga2007parallel}, with domain-specific bilingual glossaries built from \textit{IATE} dictionaries~\citelanguageresource{iate}.

The resulting parallel corpus was restricted to documents with an average \textit{document-level} alignment score above 0.3 for DE-EN and 0.5 for EN-DE.

In automatically aligned written data, empty segments typically reflect alignment errors rather than genuine translation features.
Documents containing aligned empty source (\texttt{src\_seg}) or target (\texttt{tgt\_seg}) segments were excluded, except when such segments occurred at the beginning or end of a document or contained three words or fewer; in those cases, only the segment was removed. This filtering affected 1,235 document pairs in DE-EN direction and 1,072 document pairs in EN-DE (i.e., about 10\% in each direction), most of which were removed entirely. 

A manual evaluation of 80 randomly selected document pairs, where originals had approximately the same number of segments (750 segment pairs per direction), conducted by a compensated research assistant, showed misalignment rates below 4.5\% for DE-EN and 1.8\% for EN-DE.

Given the strong manual evaluation results and the much larger size of the written subcorpus (compared to spoken), testing alternative alignment algorithms such as \textit{Vecalign}~\cite{thompson2019vecalign}, which mainly benefits low-resource cases, was deemed unnecessary.

\subsection{Text Standardisation}\label{ssec:prepro}
The same set of text preprocessing operations was applied uniformly to all texts before any word-level annotation. These operations were guided by our prior experience with the data and by the intended use of the corpus for cross-modal, cross-lingual, and monolingual comparisons, enabling consistent analysis across both translation directions.

The preprocessing steps included the removal of non-printable ASCII characters (except tabs and newlines), standardisation of single and double quotes, dashes and superscripts. All hyphenated words were force-tokenised to improve comparability across languages (e.g., \textit{EVP-Fraktion}, \textit{well-known}).

\subsection{Annotation}\label{ssec:anno}
With this corpus release, we provide three layers of word-level annotation: standard morphological and syntactic annotation, word surprisals derived from base and fine-tuned GPT-2 and MT models, and word alignments.

\paragraph{Stanza parse.} 
Linguistic annotation (tokenisation, POS tagging, lemmatisation, and dependency parsing) was performed using Stanza v1.10.1~\cite{qi2020stanza} with default pre-trained models for English and German, corresponding to the Universal Dependencies treebanks distributed with this release. Segment alignment was preserved throughout the parsing process.

Multi-sentence segments (about twice as frequent in the spoken subcorpus) were parsed sentence by sentence and retained within the same aligned translation unit by maintaining continuous word numbering in \texttt{word\_ids} across each segment.
The main difference between annotating the spoken and written subcorpora was that, for spoken data, FPs had to be removed prior to annotation and then reinserted at their original positions into the parsed segment with \texttt{FP} tag using the \texttt{word\_ids} index.
The vertical CoNLL-U output of \textit{Stanza} was used as the canonical tokenisation scheme used in the subsequent indexing with LLM-based surprisals and for word alignment, subject to an important caveat concerning \textit{multitokens} discussed below.

Multitokens are multiword tokens, most notably English contractions (\textit{it’s}, \textit{there’s}, \textit{I'd}) and Genitive case forms and syncretic German prepositions such as \textit{am} and \textit{zur}. As illustrated in Appendix A (Figure~\ref{fig:vrt}), these forms are represented simultaneously at two levels in the data. On one level, surface multiword tokens are preserved for surprisal annotation and word alignment, which operate on the actually realised forms. On the other level, the internal morphosyntactic structure of these contractions is made explicit through multitoken expansion (e.g., it's$\rightarrow$it is, im$\rightarrow$in dem, December's$\rightarrow$December 's), which is required for fully functional syntactic dependency trees. The true surprisal values exist only for surface realisations; the surprisal values in the rows containing expansions are set to N/A, while the surface token word\_id receives an additional identifier (e.g. SI\_DE\_EN\_030-21:010:1 for there in there's).

To enable comparisons across modes and translation directions, the written subcorpus was divided into \textit{train} and \textit{test} splits. Prior to this split, any speeches occurring in both written and spoken modes were removed from the written data.

The \textit{test} split consists of documents selected to match the spoken subcorpus in size and document length (at least 12 segments per document; 170 document pairs per translation direction), ensuring comparable segment counts. The spoken corpus was assigned entirely to the \textit{test} split.

The \textit{train} split was balanced across translation directions by randomly subsampling the larger German–English portion to match the English–German data in terms of segment count. This split was used to fine-tune the LLMs for surprisal estimation. The written data are released in separate tables corresponding to the train and test splits. The resulting data split parameters are presented in Table~\ref{tab:split}.

\begin{table}[]
	\resizebox{\columnwidth}{!}{%
	\begin{tabular}{@{}p{1cm}llrrr@{}}
		\toprule
        mode                & lpair              & type  & words    & segs &  docs\\ \midrule
        \multirow{4}{*}{spoken} & \multirow{2}{*}{DE-EN} & src, org & 63,172  & \multirow{2}{*}{3,249}& \multirow{2}{*}{165}\\
		&                        & tgt & 61,826 &  & \\
		& 						\multirow{2}{*}{EN-DE} & src, org & 70,130 & \multirow{2}{*}{3,440} & \multirow{2}{*}{137}\\
		&                        & tgt & 63,727  &  & \\ \midrule

		\multirow{4}{=}{\parbox{1cm}{written (test)}} & \multirow{2}{*}{DE-EN} & src, org & 76,327  & \multirow{2}{*}{3,218} & \multirow{2}{*}{170}\\
		&                        & tgt & 85,053 & & \\
		& \multirow{2}{*}{EN-DE} & src, org & 75,804  & \multirow{2}{*}{3,070}& \multirow{2}{*}{170}\\
		&                        & tgt & 75,668  &  & \\ \midrule
		
		\multirow{4}{=}{\parbox{1cm}{written (train)}} & \multirow{2}{*}{DE-EN} & src, org & 2,924,739  & \multirow{2}{*}{122,407} & \multirow{2}{*}{10,367}\\
		&                        & tgt & 3,292,542 & & \\
		& \multirow{2}{*}{EN-DE} & src, org & 3,110,670  & \multirow{2}{*}{122,233} & \multirow{2}{*}{11,690}\\
		&                        & tgt & 3,102,631  &  & \\
		\bottomrule
	\end{tabular}
	 }
	\caption{\label{tab:split}Corpus statistics for \textit{EuroParl-EPIC-UdS}, based on vertical format. Spoken data counts include FPs and empty segments on either side. Multiword tokens are counted once as surface-level items. Abbreviations: lpair=language pair, src=source, org=original, tgt=target; docs=documents, segs=segments.}
	
\end{table}

To further characterise the data, we report additional descriptive parameters. Average segment length ranges from 24.1 to 26.8 words for written data and 18.6 to 20.7 words for spoken data, with standard deviations of approximately 14 words. Minimum segment length is not controlled for and can be as short as one word. Maximum segment length reaches 317 words in the written German target, where multiple sentences are merged into a single translation segment.

In spoken data, targets contain roughly twice as many multi-sentence segments as sources (DE–EN: 8.4\% vs. 17.4\%; EN–DE: 7.8\% vs. 19.9\%). In written data, multi-sentence segments are rare overall (1.1–7.5\%), and in contrast to spoken, shows a clear directional asymmetry: English-to-German favours sentence splitting, while German-to-English favours sentence merging, implying denser information packaging in written English. For the remainder of the paper, we treat the segment—rather than the sentence—as the operational unit, although the vast majority of aligned segments contain a single sentence on either the source or the target side.

\paragraph{Surprisal indexing.}
Surprisal is a key measure of information in the information-theoretical approach to language study. It is calculated as an inverse base-2 log probability of a word in context (Equation~\ref{eq:srp}):

\begin{equation}
	S(w) = -log_2(P(w | context))
	\label{eq:srp}
\end{equation} 

\begin{figure}[!t] % [H] forces it to stay exactly HERE
\begin{flushleft} % Ensures no weird indentation at the top
\section*{Appendix C. Fine-Tuning Details}
\end{flushleft}

\begin{minipage}[t]{0.46\linewidth}
    \vspace{0pt}
    \centering
    \begin{tabular}[t]{lrr} % Added [t] here as well
    \toprule
    Hyperparameter & GPT-2 & MT \\
    \midrule
    batch\_size & 32 & 16 \\
    gradient\_accumulation\_steps & 2 & 3 \\
    warmup\_ratio & 0.05 & 0.10 \\
    epochs & 10 & 10 \\
    learning\_rate & $8 \times 10^{-6}$ & $8 \times 10^{-6}$ \\
    weight\_decay & 0.01 & 0.001 \\
    early\_stopping $\delta$ & 0.0005 & 0.001 \\
    early\_stopping\_patience & 3 & 3 \\
    max\_sequence\_length & 150 & 150 \\
    label\_smoothing & 0.0 & 0.1 \\
    deterministic\_training & False & False \\
    \bottomrule
    \end{tabular}
\end{minipage}
\hfill
\begin{minipage}[t]{0.48\linewidth}
    \vspace{0pt}
    \centering
    \includegraphics[width=\linewidth]{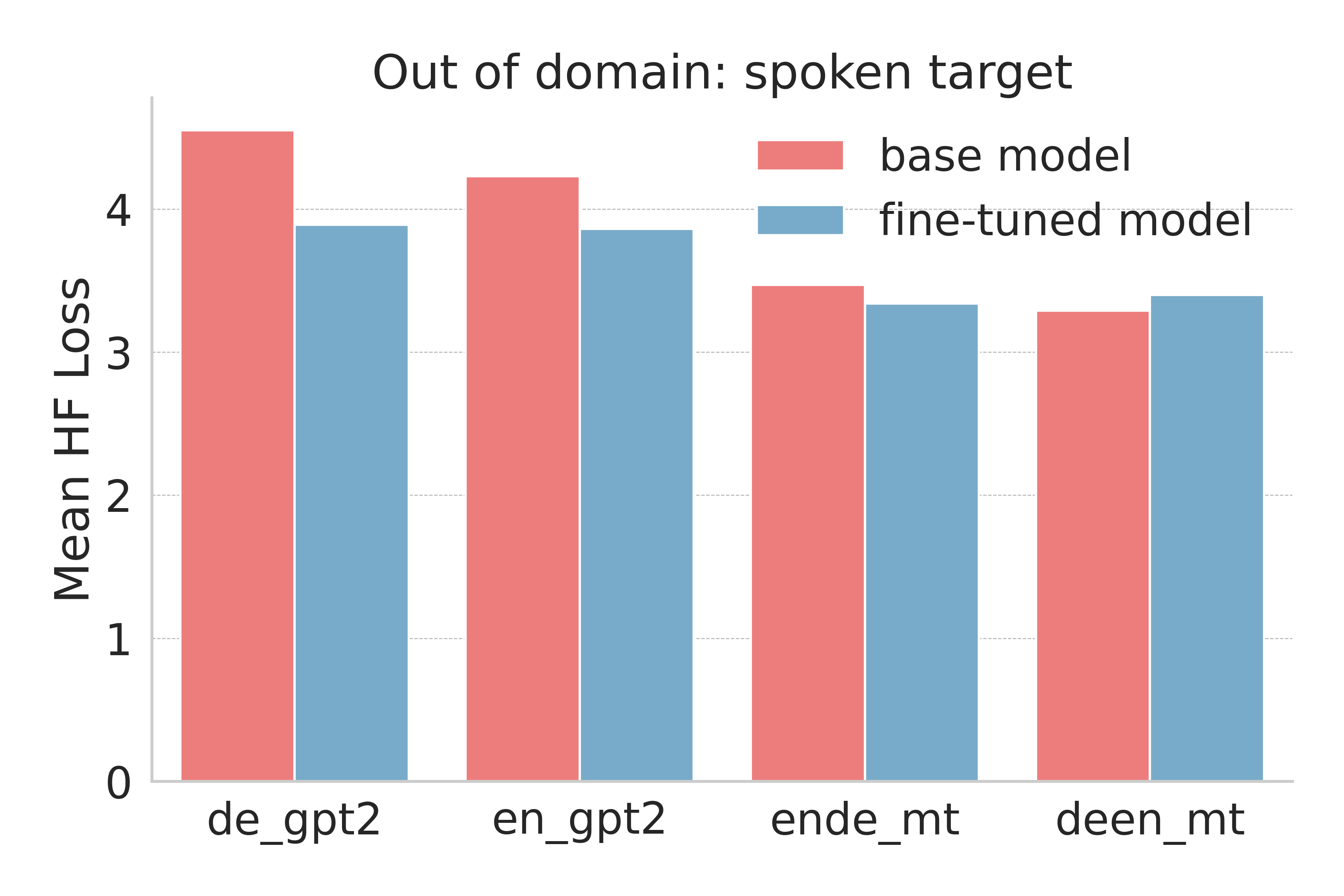}
\end{minipage}

\vspace{1em}

\begin{minipage}[t]{0.46\linewidth}
    \captionof{table}{\label{tab:hyperpars}Training hyperparameters for GPT-2 and MT systems.}
\end{minipage}
\hfill
\begin{minipage}[t]{0.46\linewidth}
    \captionof{figure}{\label{fig:bars}Hugging Face cross-entropy loss on out-of-domain test sets.}
\end{minipage}
\end{figure}

\begin{figure*}[!t]
    \centering
    \begin{subfigure}[b]{0.48\linewidth}
        \centering
        \includegraphics[width=\linewidth]{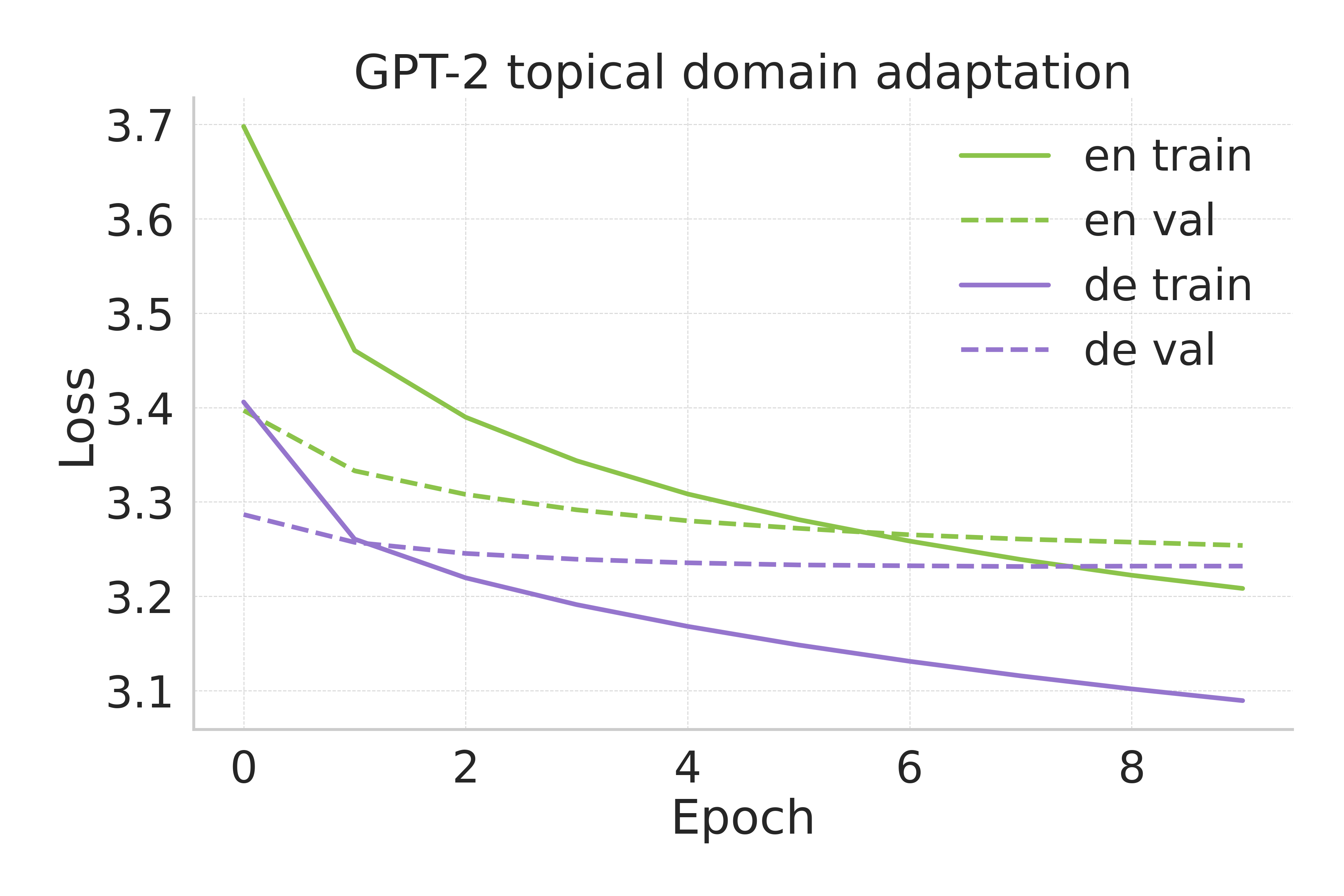}
        \caption{GPT-2 losses}
        \label{fig:gpt2_losses}
    \end{subfigure}
    \hfill
    \begin{subfigure}[b]{0.46\linewidth}
        \centering
        \includegraphics[width=\linewidth]{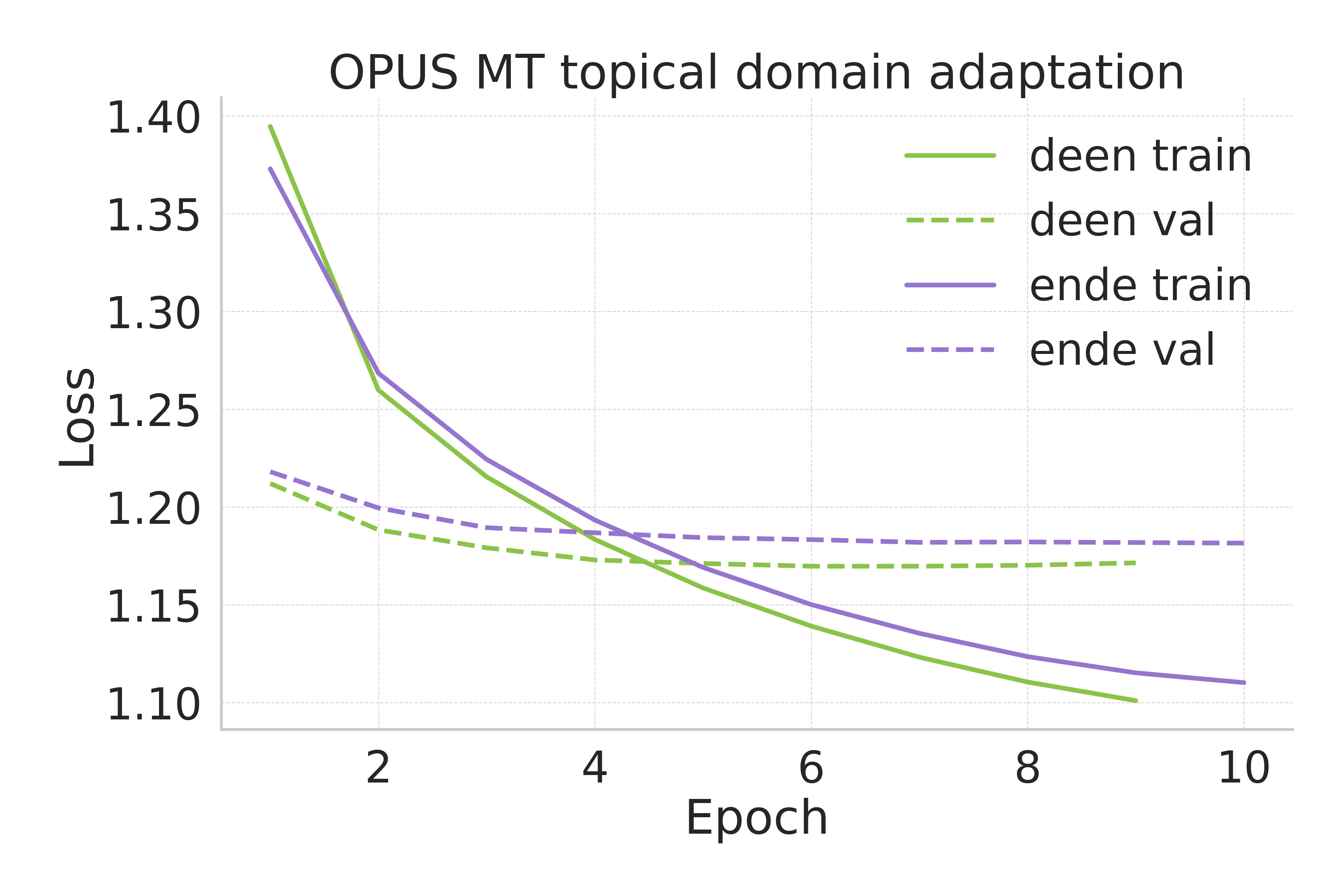}
        \caption{MT losses}
        \label{fig:mt_losses}
    \end{subfigure}

    \caption{\label{fig:tune_tracking}Training and validation loss curves for finetuning experiments. (a) GPT-2 base model, (b) MT baseline. The curves have a slow adaptation profile due to a conservative learning rate ($8 \times 10^{-6}$) and weight decay. While validation progress is gradual, this parameterisation was necessary to prevent overfitting.}
\end{figure*}

\begin{figure*}[ht]
    \includegraphics[width=\linewidth]{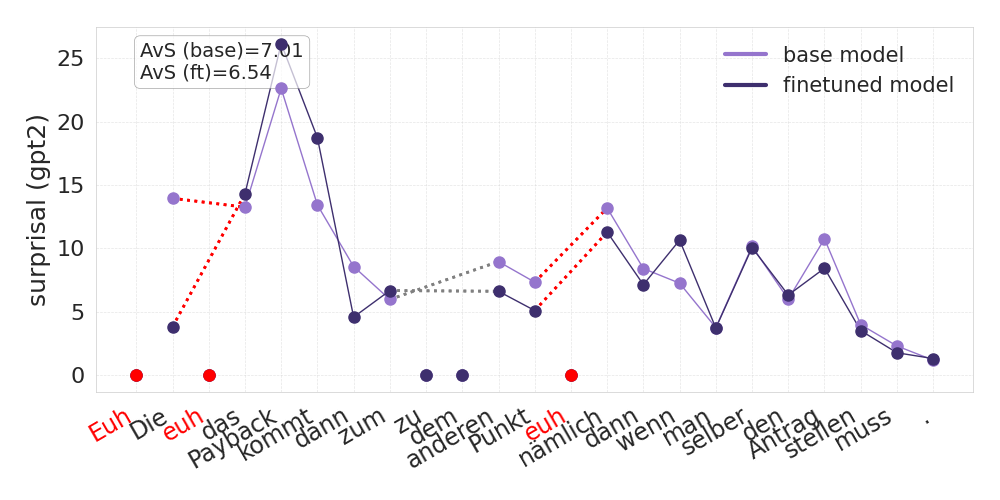}
    \caption{\label{fig:base_vs_ft_gpt2_case}Word-surprisals from base and fine-tuned GPT-2 models for a target with filler particles and expanded UD multiwords (seg\_id: SI\_EN\_DE\_129-27). Domain-specific vocabulary (e.g. \textit{Antrag}) is less surprising for fine-tuned models, while general language words (e.g. \textit{kommt}) are more surprising compared to base models.}
\end{figure*}

The probabilistic measures in this study are derived from monolingual GPT-2 small models and neural MT models. The GPT-2 small family was selected because its surprisal estimates have been shown to correlate best with experimental measures of processing difficulty~\cite{oh2023}.
Each word (defined as a Stanza token in vertical corpus format) in the EPIC-EuroParl-UdS test subsets is annotated with surprisal values obtained from language-specific pre-trained (base) and fine-tuned GPT-2 models 
(English~\citet{radford2019language}; German~\citetlanguageresource{GPT2-German}). 
In addition, each target word is assigned surprisal values derived from pre-trained and fine-tuned, translation-direction-specific MT models, developed by~\citet{tiedemann-thottingal-2020-opus}. The vocabulary size of all four models is approximately 50 K tokens.
Fine-tuning was conducted exclusively on the written train split of the corpus (comprising written originals and written parallel data for the MT baseline). Consequently, the spoken targets represent an out-of-domain (OOD) challenge for the MT models. For the GPT-2 models, this challenge is twofold: the targets are OOD in terms of both mode (spoken vs. written) and text type (original vs. targets). Detailed training hyperparameters are documented in Table~\ref{tab:hyperpars} (Appendix C), followed by a comprehensive evaluation of the resulting fine-tuned models. The training process yielded expected results: the cross-entropy loss for fine-tuned models is lower than for base models even on out-of-domain test sets (except deen\_mt, see Figure~\ref{fig:bars}); the training and validation losses go down with no signs of overfitting (Figure~\ref{fig:tune_tracking}); subcorpus-averaged surprisal and BLEU is better for fine-tuned than for base models (Table~\ref{tab:avs}) and individual examples demonstrate that domain specific vocabulary is less surprising for fine-tuned models (Figure~\ref{fig:base_vs_ft_gpt2_case}).
Train subsets are included in the corpus and are indexed with surprisal from base models only.

\begin{figure*}[!t]
\centering
\section*{Appendix B. Sliding Window Results}
\vspace{0.5em}
    \includegraphics[width=\linewidth]{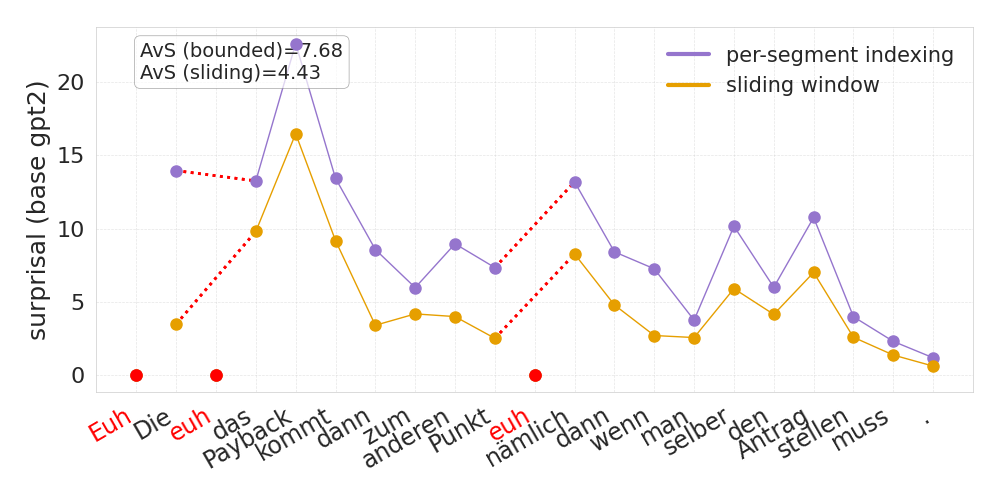}
    \caption{\label{fig:gpt2_approaches_case} Word surprisals from base GPT-2 using segment-bounded and sliding window approaches (seg\_id: SI\_EN\_DE\_129-27)}
\end{figure*}

\begin{figure*}[!t]
    \includegraphics[width=\linewidth]{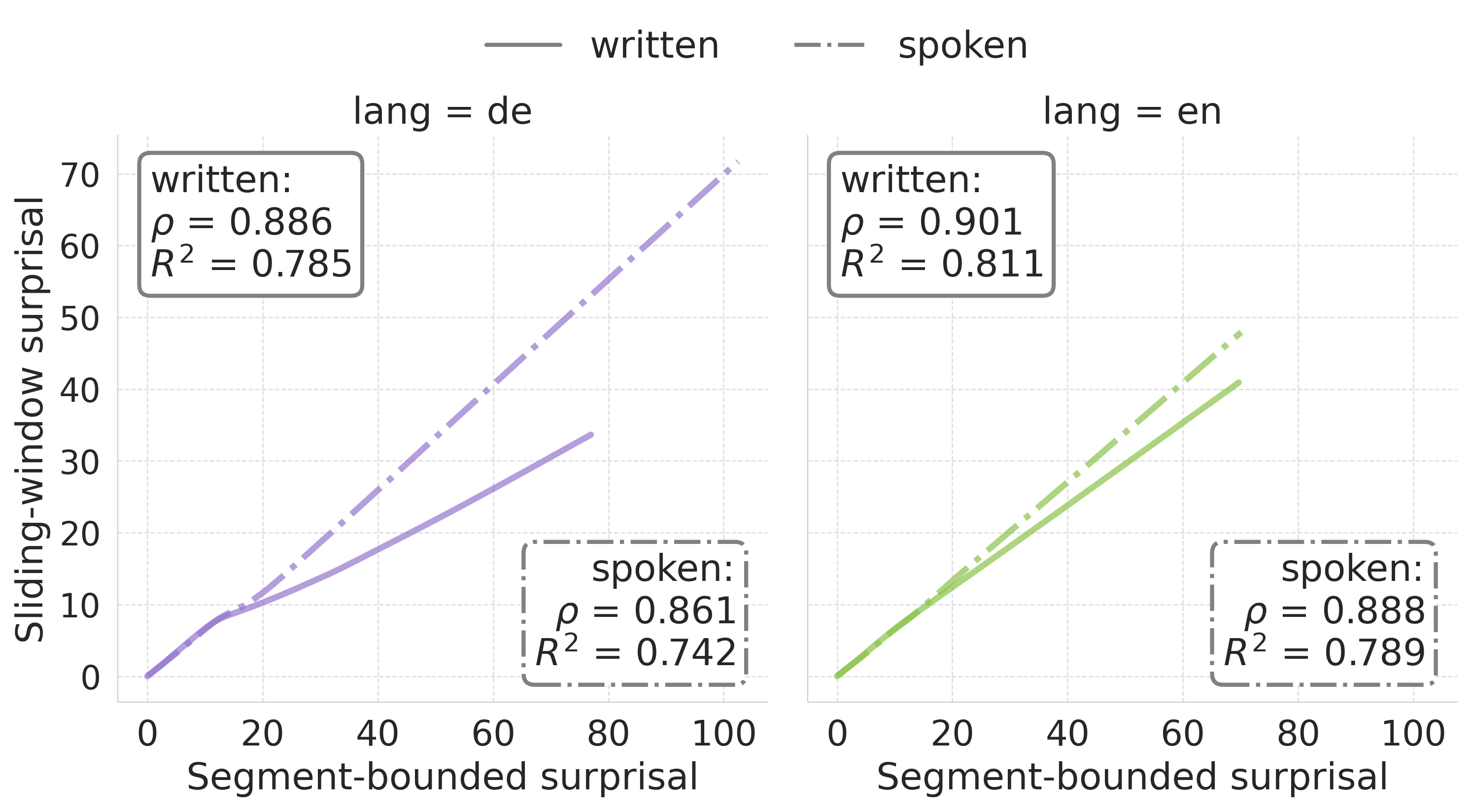}
    \caption{\label{fig:gpt2_approaches} Relation between segment-bounded and sliding window base GPT-2 word surprisals for written and spoken target subcorpora by language. Lines represent LOWESS-smoothed fits (frac = 0.2).}
\end{figure*}

Word-level surprisal is computed within segment boundaries. For each subword, the probability is conditioned on the left context up to the segment boundary, explicitly marked with \textit{<|bos|>} token. Rare, longer segments in the training dataset are truncated to 150 leftmost subwords. Consequently, surprisal is estimated with a varying-length left context, capped at 150 tokens. This setup was compared with a 64-subword sliding-window approach. The two methods were strongly correlated, yielding linear regression $R^2$ between 0.74 (spoken German targets) and 0.81 (written English targets) (see Figure~\ref{fig:gpt2_approaches}, Appendix B). Because the sliding-window approach produced comparable patterns, albeit on a reduced scale (as illustrated in Figure~\ref{fig:gpt2_approaches_case}, Appendix B), it was not pursued further, and the segment-bounded model was adopted for all subsequent analyses.

Our approach to input formatting and post-processing was guided by two principles: maintaining the naturalistic quality of the source text and ensuring a robust alignment between subword-level surprisals and linguistic tokens.

Unlike standard pipelines that rely on expanded syntactic representations, we opted for naturalistic raw text input. This involved detokenising punctuation, preserving hyphenated tokens as single units, and utilising surface multiword tokens (e.g., zur, he’ll, Council’s) rather than their expanded Universal Dependencies (UD) forms (e.g., zu der, he will, Council 's). This was critical for accuracy: preliminary tests showed that feeding fully tokenised segments inflated probabilities for punctuation and multiword tokens, as the underlying language models were pre-trained on raw text.

LLM subword probabilities were converted to log space and pre-aggregated based on the model’s beginning-of-word markers (e.g., GPT-2's \textit{Ġ}), while keeping punctuation as separate units. However, this Ġ-based pre-aggregation did not guarantee an exact match with Stanza tokens. In approximately 3\% of written test segments, mismatches occurred where a single pre-aggregated unit spanned multiple Stanza tokens (e.g., Stanza: ['über', '99', '\%', '.']; GPT-2: ['Ã¼ber', '99', '\%.']) or vice versa (Stanza: ['at', '5.30', 'p.m.', 'on', 'Thursday']; GPT-2: ['at', '5.30', 'p.m', '.', 'on', 'Thursday']). These cases appeared almost exclusively around complex punctuation sequences. 
Rather than discarding these segments and compromising the dataset, we implemented a sequential set of rules to recover values for common edge cases, including abbreviations (i.e., p.m.), float-like numbers (20 000, 0,7\%-Zielgröße), and punctuation sequences ('\%,', 'IV.,', '):'). When subword values required splitting to align with Stanza tokens, 0.75 × value was assigned to the preceding token and 0.25 × value to the trailing punctuation mark; alternatively, subword surprisals were summed.

Despite our efforts to recover most erroneous segments, a few dozen cases in the 120K-segment train splits remained misaligned. Rather than excluding these segments, we indexed them with N/A values. This preserves the original segment alignment and document structure, facilitating future document-level analyses and the annotation of higher-level discourse phenomena, such as coreference, without sequence gaps.

The vertical format of the spoken subcorpus retains FPs and empty segments to preserve the characteristics of spoken language. 
However, these elements were removed when computing surprisal values.
They were reinserted during post-processing at their original positions, with all corresponding surprisal values set to N/A.

\begin{figure*}[!ht]
        \centering
        \includegraphics[width=\linewidth]{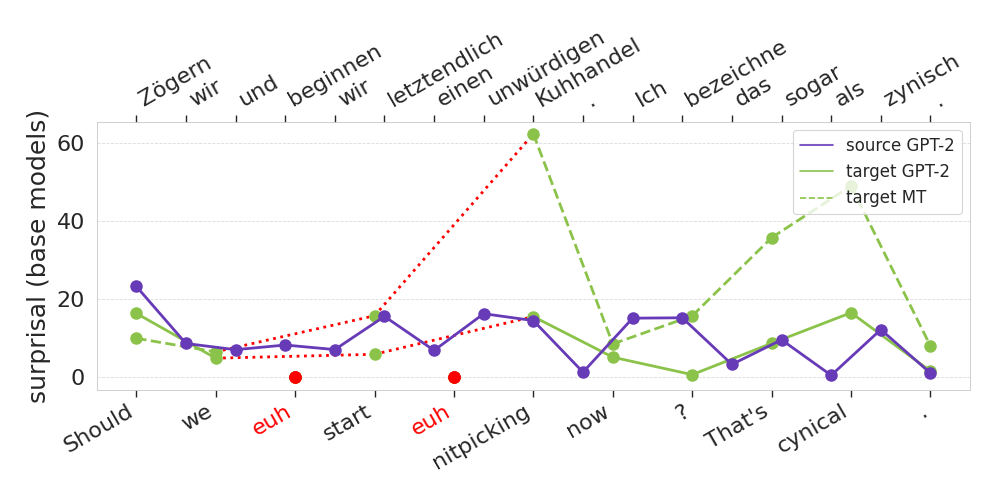}
        \caption{\label{fig:word_srp}Word surprisals from base monolingual GPT-2 models and base MT model for a segment pair.}
\end{figure*}

Segment-level corpus formats contain average segment surprisal (AvS), computed both at the Stanza token level (excluding FPs) and at the subword level (e.g., for GPT-2 columns: \texttt{base\_gpt\_AvS}, \texttt{base\_gpt\_AvS\_subw}, \texttt{ft\_gpt\_AvS}, \texttt{ft\_gpt\_AvS\_subw}). Additionally, we included segment-level \textit{pseudo-BLEU} scores (\texttt{base\_bleu}, \texttt{ft\_bleu}) calculated between the reference target segment and token-wise argmax predictions under gold target prefixes, using \textit{sacrebleu}~\cite{post-2018-call}.
To avoid overly harsh zero scores when higher-order n-grams do not match, we apply exponential smoothing (smooth\_method="exp"). This approach allows sentence-level BLEU to reflect partial literal matches, making it more stable and informative than strict n-gram matching.
Higher BLEU scores indicate that the MT model finds the translation more straightforward or typical, suggesting lower cross-lingual transfer difficulty.

Figure~\ref{fig:word_srp} shows a spoken parallel segment containing FPs and two sentences (seg\_id: \_DE\_EN\_023-08). It reports three types of word-level surprisal values: base monolingual GPT-2 surprisals for source and target words, and base cross-lingual MT surprisals for target words.  %, which conditions the probabilities in the target on the source segment and preceding target words (if any) as context. 

Table~\ref{tab:avs} (Appendix B) offers subcorpus-averaged surprisals for each text type and model (both base and fine-tuned). 
While these averages show only small differences between categories, word-level surprisal captures local phenomena (e.g. FPs), justifying the release of word-level surprisals in this corpus.

A key finding is the divergent behaviour of the monolingual and MT models after fine-tuning. For GPT-2, fine-tuning consistently reduces average surprisal across all subcorpora. Conversely, MT models exhibit higher surprisal after fine-tuning, despite a concurrent increase in pseudo-BLEU scores (indicating better translation quality). Furthermore, mean surprisals are systematically higher for the spoken mode compared to the written mode, likely reflecting the higher entropy and structural irregularities inherent in spoken discourse.

The relationship between GPT-2 and MT indices is considered to capture the trade-off between accuracy and fluency objectives in translation~\cite{teichTranslationInformationTheory2020}. Theoretically, translating requires balancing fidelity to the source (low MT surprisal) with fluency in the target (low GPT-2 surprisal). When this balance is hard to achieve, translators can opt for a literal translation strategy (low MT surprisal) at the expense of target language fluency (high GPT-2 surprisal) or deviate from the source (high MT surprisal) in order to comply with target norms (low GPT-2 surprisal). Based on this reasoning, ~\citep{lim-etal-2024-predicting} expected a negative correlation between MT and GPT-2 surprisal. Their results, computed over the combined data from all 13 translation directions using multilingual GPT-2 and MT surprisals, confirm the hypothesis. They report a single weak but statistically significant negative correlation between MT and GPT-2 surprisal. 

Our data, however, reveals a consistent nonlinear relation between GPT-2 and MT surprisals across subcorpora. The hypothesised negative correlation holds only for simpler segments (below 11 bits/word). 
Above this threshold, increased deviation from the source do not contribute to target fluency, quite the opposite.

Generalised Additive Models (GAM) confirm that these nonlinearities are systematic and predictive ($R^2$ 0.17–0.28), suggesting that the hypothesised accuracy-fluency trade-off collapses as cross-lingual transfer difficulty increases (see Figure~\ref{fig:gam} in Appendix C). 

The differences in findings likely stem both from the neural models used for surprisal estimation and from the nature of the data. Unlike prior work based on small-scale, experimentally elicited student translations, our study draws on professionally produced, high-quality translations in a comparatively more demanding domain, which may fundamentally alter the observed accuracy–fluency dynamics. We further argue that the degree of nonlinearity in cross-lingual transfer captured by MT surprisal reflects the scale of production effort required by a given segment.
Once this effort exceeds a certain threshold, neither translators nor interpreters can consistently maintain audience-oriented fluency, pointing to inherent constraints of the production process itself.%  and for very challenging segments (average MT surprisal > 20, EN-DE only)

\paragraph{Word alignment.}
Inspired by AWESoME~\cite{dou2021word}, we use the \textit{bert-base-multilingual-cased} model~\cite{BERTbasemultilingualcased} to extract contextualised subword representations. For every pair of source and target subword embeddings, we compute dot-product similarity and apply bidirectional softmax normalisation (source$\rightarrow$target and target$\rightarrow$source). Subword pairs whose mutual bidirectional softmax score exceeds the recommended threshold of 0.01 are marked as aligned; this value constitutes the alignment confidence threshold used throughout the paper.

Aligned subword pairs are then aggregated to the word level using the pre-computed subword–word maps. For each word pair, the final alignment score is obtained as the mean of the bidirectional softmax scores, averaged across all contributing subword pairs. If a source word exceeds the same alignment confidence threshold at the word level, its corresponding target word(s) are recorded in the columns \texttt{aligned\_word} and \texttt{aligned\_word\_id}.

\begin{table}[]
\resizebox{\columnwidth}{!}{%
\begin{tabular}{@{}llrrr@{}}
\toprule
% updated 10 Feb 2026
mode    & lpair & src\_tokens & \% unaligned & \% multi-aligned \\ \midrule
spoken  & DE-EN  & 60,125              & 34.96      & 4.42            \\
spoken  & EN-DE  & 66,729              & 34.48      & 2.14             \\
written & DE-EN  & 76,327              & 19.59      & 5.08            \\
written & EN-DE  & 75.804              & 20.03      & 1.49             \\ \bottomrule
\end{tabular}
}
\caption{\label{tab:word_align}Word alignment statistics for test sets source tokens. Unaligned and multi-aligned columns have ratios of source tokens not aligned or aligned to several target tokens.}
\end{table}

The softmax-based alignment permits one-to-many mappings; in such cases, these columns contain lists of target words and their IDs. As shown in Table~\ref{tab:word_align}, this word alignment pattern is direction-dependent, occurring in over 4\% of source words in the DE–EN data and in about 2\% in EN–DE, irrespective of mode. In contrast, the proportion of unaligned source words (marked with an explicit `NA' string in the data tables) reflects mode-specific variation rather than language-pair differences. As expected, the spoken subcorpora exhibit a higher proportion of unaligned source words (up to 35\%), consistent with the observation that interpreting tends to yield less literal, less isomorphic renditions.

\section{Applications of Information Theory}\label{sec:applications}

This section presents possible applications of the corpus. While section \ref{ssec:applications1} outlines a novel study, section \ref{ssec:applications2} refers to a related study and describes ideas for further possible applications. 

\subsection{Interpreting: FP Prediction}\label{ssec:applications1}
%\subsection{Example applications: surprisal}
In this study, we look at the link between FPs and surprisal. Previous work has shown that FPs are followed by high-surprisal words, ~\cite{zamevcnik2019,dammalapati2021, raoof2024bilingualsrpfiller}, indicating upcoming lexical retrieval difficulty. Our previous study \cite{pollklaesener2025surprisal} replicated this effect in interpreting. However, we focused solely on the effect of target language surprisal (formulation difficulty). We now extend this work using the new annotation to include source surprisal (comprehension difficulty), MT surprisal (transfer difficulty), word alignment, and a comparison of base and fine-tuned models.

Using mixed-effects logistic regression, we model if a target word is preceded by a FP (1) or not (0). FPs make up roughly 5\% of all target words in both EN–DE and DE–EN (3,255 and 2,328 instances, respectively). Predictors represent different types of processing difficulty: source surprisal (comprehension), target surprisal (formulation), and MT surprisal (transfer). Both local (word-level) and global (segment-average) measures of the three surprisal types are included in the model as global surprisal contributes explanatory value \citep{pollklaesener2025euh}. Every target word in the corpus is annotated with target and MT surprisal. To obtain comprehension surprisal, we use word- and segment-alignment. Multi-aligned target words are averaged and unaligned target words are excluded. Separate models are estimated for EN–DE and DE–EN directions as English and German surprisal values are not directly comparable.

First, we compare fine-tuned and base surprisal. To do so, we fit models using either base or fine-tuned surprisal values for each direction. In both directions, models with base surprisal have lower AIC values than fine-tuned surprisal models (DE-EN: base AIC = 11359, fine-tuned AIC = 11473; EN-DE: base AIC = 15313, fine-tuned AIC = 15349), indicating better model fit. Similarly, models with base surprisal achieve higher C-scores, reflecting better discrimination between outcomes (DE-EN: base C = 0.6994, fine-tuned C = 0.686; EN-DE: base C = 0.6515, fine-tuned C = 0.6432). We conclude that base surprisal values perform better in this task.

\begin{table}[h!]
\centering
\small
\begin{tabular}{@{}l|>{\raggedleft\arraybackslash}c>{\raggedleft\arraybackslash}c|
                    >{\raggedleft\arraybackslash}c>{\raggedleft\arraybackslash}c@{}}
\toprule
 & \multicolumn{2}{c|}{\textbf{DE-EN}} & \multicolumn{2}{c}{\textbf{EN-DE}} \\ 
 & Coef & SE & Coef & SE \\
\midrule
Intercept     & -3.49* & 0.18 & -3.04* & 0.08 \\
nxtwS\_tgt     &  0.54* & 0.03 &  0.37* & 0.03 \\
nxtwS\_src     & -0.10* & 0.03 & -0.14* & 0.03 \\
nxtwS\_mt      &  0.04  & 0.03 &  0.08* & 0.02 \\
AvS\_tgt       & -0.35* & 0.04 & -0.40* & 0.03 \\
AvS\_src       &  0.17* & 0.03 &  0.20* & 0.03 \\
AvS\_mt        & -0.20* & 0.03 & -0.03  & 0.03 \\
\bottomrule
\end{tabular}
\caption{Fixed-effect estimates (Coef) and standard errors (SE) from mixed-effects logistic regression models predicting FPs in the target. Surprisal values are from base models. Predictors are z-scored. Asterisk indicates statistical significance (p<.05).}
\label{tab:FP_study1}
\end{table}

Table \ref{tab:FP_study1} presents the fixed-effect estimates for the models using base surprisal values. Given the consistency of the results across directions, we present a joint discussion below. Formulation surprisal measures (nxtwS\_tgt, AvS\_tgt) emerge as the strongest predictors for FP occurrence, which is noteworthy, as we had expected transfer difficulty to be the primary indicator of production difficulty in interpreting. One possible explanation is that MT surprisal is computed using more source context (the entire aligned source segment) than the interpreter may have heard at a given moment. Future studies could take into account time alignment and estimate surprisal based only on the source context available to the interpreter up to the last word heard.

Interestingly, the coefficient for average target surprisal is negative, indicating that lower average segment surprisal is associated with higher FP probability. This contrasts with the findings reported in \citet{pollklaesener2025euh}, though considering all three surprisal measures together may clarify this pattern. For next-word surprisal, formulation and transfer difficulty (nxtwS\_tgt, nxtwS\_mt) show positive effects, whereas comprehension difficulty (nxtwS\_src) has a negative effect. Interpreters thus tend to produce FPs before words that are difficult to formulate and transfer but easy to comprehend. We see the opposite for the global measures. Average target and MT surprisal (AvS\_tgt, AvS\_mt) show negative effects, while average source surprisal (AvS\_src) has a positive effect on FP occurrence. 
This pattern suggests that comprehension difficulties may not surface at the level of the immediately produced word. Instead, they appear to manifest more globally, with processing demands accumulating over the segment. Moreover, interpreters appear to compensate for the high cognitive load of retrieving the next word by producing material with low transfer and formulation difficulty in the rest of the segment.

\subsection{Cross-Modal Studies}\label{ssec:applications2}

In~\citet{Kunilovskaya2026subs}, EPIC-EuroParl-UdS is used to examine translationese. The study asks whether translation task difficulty can explain the degree to which translations deviate from comparable original texts in the target language, assuming that producing translationese- or interpretese-ridden text requires less effort. Translationese is quantified as the segment-level probability of being a translation (\textit{translatedness} score) returned by a binary classifier. This score as the response variable in a regression experiment. Translation task difficulty is separated into comprehension difficulty, operationalised by a set of source-related predictors such as source text surprisal or lexical density, and transfer difficulty, operationalised by predictors reflecting the relationship between source and target, such as MT surprisal or mean alignment score.  
The results show that transfer difficulty is more effective in predicting translatedness than comprehension difficulty (except in spoken DE-EN). Information-theoretic predictors (e.g. surprisal, entropy) match or outperform traditional features (e.g. tree depth, lexical density) in written mode, but offer no advantage in spoken mode. 

EPIC-EuroParl-UdS can also be used to examine the influence of V-final clauses in German sources on English spoken and written targets. While translators can access the entire source text before producing their output, interpreters render segments in real time, often having to begin their interpretation before the sentence has fully unfolded. \citet{hodzik2017predictive} investigated the prediction of sentence-final verbs in simultaneous interpreting form German into English. They found that more predictable verbs (as indicated by contextual semantic cues) were associated with shorter latency times, suggesting facilitated production through predictability. 
This effect could be investigated using EPIC-EuroParl-UdS, relating the surprisal of the clause-final verbs in the German source to the surprisal of the aligned target words to reveal transfer strategies in translation and interpreting.

\section{Conclusion}
This paper presents the surprisal-annotated, updated English$\leftrightarrow$German translation and interpreting corpus \textit{EPIC-EuroParl-UdS} and its possible applications. We correct and refine the corpus content and structure, and introduce new annotation layers designed for information-theory-motivated studies, especially cross-lingual ones, at the local level. The corpus pre-processing pipeline emphasises document integrity by resolving issues with individual segments instead of skipping them. 

By providing this resource, we hope to facilitate advances in contrastive and translationese studies, especially since the vertical format is directly openable in \textit{R}, simplifying direct data analysis. All data and code are available under a creative commons licence. We plan to further improve \textit{EPIC-EuroParl-UdS} by incorporating surprisal measures from other models (e.g., multilingual GPT-2 and LLaMA) and adding time-alignment. 

\section{Acknowledgements}
This research was supported by the Deutsche Forschungsgemeinschaft (DFG, German Research Foundation) -- SFB 1102 Information Density and Linguistic Encoding, Project-ID 232722074.

\section{Bibliographical References}\label{sec:refs}

%\bibliographystyle{lrec2026-natbib}
%\bibliography{lrec2026}
%\nocitelanguageresource{EuroParl-UdS, EPIC-UdS, OPUS-MT-DE-EN, OPUS-MT-EN-DE, GPT2-English, bert-base-multilingual-cased, stanza, iate}

\section{Language Resource References}
\label{sec:lr_ref}

%\bibliographystylelanguageresource{lrec2026-natbib}
%\bibliographylanguageresource{languageresource}

\clearpage
\onecolumn
\begin{figure*}[!t]
\centering
    \section*{Appendix A. Details on Corpus Formats}
\begin{minipage}[t]{0.48\linewidth}
\begin{verbatim}
--spoken
 |--epic-clean_long.tsv.gz
 |--epic-fp-nones_long.tsv.gz
 |--epic-clean_wide.tsv.gz
 |--epic-fp-nones_wide.tsv.gz
 |--epic-clean_vrt.tsv.gz
 |--epic-fp-nones_vrt.tsv.gz
 |--epic_org_doc_meta.tsv.gz
\end{verbatim}
\end{minipage}%
\hfill
\begin{minipage}[t]{0.48\linewidth}
\begin{verbatim}
--written
 |--europ-test_long.tsv.gz
 |--europ-train_ende_long.tsv.gz
 |--europ-train_deen_long.tsv.gz
 |--europ-test_wide.tsv.gz
 |--europ-train_ende_wide.tsv.gz
 |--europ-train_deen_wide.tsv.gz
 |--europ-test_vrt.tsv.gz
 |--europ-train_ende_vrt.tsv.gz
 |--europ-train_deen_vrt.tsv.gz
 |--europ_org_doc_meta.tsv.gz
\end{verbatim}
\end{minipage}
\caption{\label{fig:structure}Structural overview of the EPIC-EuroParl-UdS repository, illustrating the organisation of spoken and written sub-corpora across data formats (long, wide, and vertical) and associated metadata.}
\end{figure*}

\begin{table*}[!t]
\centering
\small

\begin{tabular}{p{2.3cm} p{3.5cm} p{8.5cm}}
\toprule
\textbf{Format} & \textbf{Field groups} & \textbf{Fields} \\
\midrule

\textbf{Vertical} \\ (word level)
&
Modelling factors
&
doc\_id, seg\_id, word\_id, lpair, lang, mode, ttype, speaker\_id \\

&
CoNLL-U
&
id, token, lemma, pos, xpos, rel, feats, head\_id, deps, misc \\

&
Surprisal indices
&
srp\_base\_gpt2, srp\_base\_mt, \newline
srp\_ft\_gpt2, srp\_ft\_mt (except training sets)\\

&
Word alignment
&
aligned\_word\_id, aligned\_word \\

&
Context
&
raw\_seg (except training sets) \\

\midrule

\textbf{Long} \\ (segment level)
&
Modelling factors
&
doc\_id, seg\_id, lpair, lang, mode, ttype, speaker\_id, \newline
\textit{spoken}: delivery\_rate, delivery\_wpm, speech\_timing\_sec, source\_text\_delivery\_type \\

&
GPT-2 surprisal indices
&
base\_gpt\_AvS, base\_gpt\_AvS\_subw, \newline
ft\_gpt\_AvS, ft\_gpt\_AvS\_subw (except training sets)\\

&
Responses (spoken)
&
disfluencies, fillers, fillers+3 \\

&
Context
&
raw\_seg, tokens, wc\_tok \\

\midrule

\textbf{Wide} \\ (segment-pair)
&
Core identifiers
&
src\_doc\_id, src\_seg\_id, lpair, mode\\

&
Text fields
&
src\_raw\_seg, tgt\_raw\_seg 
 \\

&
MT surprisal and BLEU
&
base\_mt\_AvS\_subw, base\_mt\_AvS, base\_bleu, \newline
ft\_mt\_AvS\_subw, ft\_mt\_AvS, ft\_bleu (except training sets)\\

\bottomrule
\end{tabular}
\caption{\label{tab:fields}Data fields available in each release format. All formats are provided as gzipped UTF-8 tab-separated tables (.tsv.gz).}
\end{table*}

\begin{table*}[]
\centering
\relsize{-1}
\resizebox{\columnwidth}{!}{%
\begin{tabular}{lrrrrrrrrr}
\toprule
word\_id & id & head\_id & token & 
srp\_base\_gpt2 & srp\_ft\_gpt2 & srp\_base\_mt & srp\_ft\_mt &
aligned\_word & speaker\_id \\
\midrule
…\_030-21:001   & \textless NA\textgreater & \textless NA\textgreater & It's           &  13   & 12.5 & 35.3 & 38.4 & Begonnen, ist      & fEN3  \\
…\_030-21:001:1 & 1  & 5 & It             &  \textless NA\textgreater & \textless NA\textgreater & \textless NA\textgreater & \textless NA\textgreater & \textless NA\textgreater & fEN3  \\
…\_030-21:001:2 & 2  & 5 & 's             &  \textless NA\textgreater & \textless NA\textgreater & \textless NA\textgreater & \textless NA\textgreater & \textless NA\textgreater & fEN3  \\
…\_030-21:002   & 3  & 5 & all            &  5.9  & 6.4 & 15.8 & 19.7 & \textless NA\textgreater & fEN3  \\
…\_030-21:003   & \textless NA\textgreater & \textless NA\textgreater & euh &  \textless NA\textgreater & \textless NA\textgreater & \textless NA\textgreater & \textless NA\textgreater & \textless NA\textgreater & fEN3  \\
…\_030-21:004   & \textless NA\textgreater & \textless NA\textgreater & hm  &  \textless NA\textgreater & \textless NA\textgreater & \textless NA\textgreater & \textless NA\textgreater & \textless NA\textgreater & fEN3  \\
…\_030-21:005   & \textless NA\textgreater & \textless NA\textgreater & euh &  \textless NA\textgreater & \textless NA\textgreater & \textless NA\textgreater & \textless NA\textgreater & \textless NA\textgreater & fEN3  \\
…\_030-21:006   & 4  & 5 & very           &  6.5  & 1.6 & 7.6  & 7.5  & guten              & fEN3  \\
…\_030-21:007   & 5  & 0 & well-intended  &  16.7 & 19.5 & 45.8 & 46.4 & guten, Absichten   & fEN3  \\
…\_030-21:008   & 6  & 5 & .              &  3    & 3.4 & 10.2 & 9    & \textless NA\textgreater & fEN3  \\
…\_030-21:009   & 1  & 3 & But            &  2.2  & 1.3 & 9.9  & 15.9 & leider             & fEN3  \\
…\_030-21:010   & \textless NA\textgreater & \textless NA\textgreater & there's  & 5.6  & 10.7 & 43   & 52.1 & \textless NA\textgreater & fEN3  \\
…\_030-21:010:1 & 2  & 3 & there          & \textless NA\textgreater & \textless NA\textgreater & \textless NA\textgreater & \textless NA\textgreater & \textless NA\textgreater & fEN3  \\
…\_030-21:010:2 & 3  & 0 & 's             & \textless NA\textgreater & \textless NA\textgreater & \textless NA\textgreater & \textless NA\textgreater & \textless NA\textgreater & fEN3  \\
\bottomrule
\end{tabular}
}
\caption{\label{fig:vrt}Excerpt from the vertical-format data illustrating the handling of multiword tokens (e.g. \textit{it’s, there’s}), where surface forms used for alignment and surprisal annotation coexist with expanded morphosyntactic representations required for syntactic parsing (seg\_id: SI\_DE\_EN\_030-21). Surprisal values are rounded to 0.1. Other UD and metadata fields are omitted.}

\end{table*}

\vfill

\clearpage

\clearpage
\onecolumn

\begin{table*}[!t]
    \centering
\begin{tabular}{@{}ll|l|rr|rr|rr@{}}
\toprule
    &       & & \multicolumn{4}{c|}{mean UD token surprisal} & \multicolumn{2}{c}{subcorpus BLEU}\\
\midrule
mode & lpair & ttype & base\_gpt2 &  ft\_gpt2 & base\_mt & ft\_mt & base\_bleu & ft\_bleu \\
\midrule
\multirow{4}{*}{spoken} 
  & \multirow{2}{*}{deen} & source & 6.997 & 5.929 & -- & -- & -- & -- \\
  &                        & target & 6.419 & 5.863 & 12.195 & 13.641 & 12.508 & 12.826\\
  & \multirow{2}{*}{ende} & source & 6.488 & 5.804 & -- & -- & -- & -- \\
  &                        & target & 7.137 & 6.102 & 12.480 & 13.819 & 11.064 & 12.450\\
\midrule
\multirow{4}{*}{written} 
  & \multirow{2}{*}{deen} & source & 6.222 & 5.185 & -- & -- & -- & --\\
  &                        & target & 5.949 & 4.951 & 11.099 & 12.348 & 27.146 & 29.550\\
  & \multirow{2}{*}{ende} & source & 6.051 & 4.899 & -- & -- & -- & --\\
  &                        & target & 5.991 &  5.203 & 12.728 & 14.116 & 23.038 & 28.177\\
\bottomrule
\end{tabular}
    \caption{\label{tab:avs}Base vs. fine-tuned surprisal and pseudo-BLEU scores for EuroParl-EPIC-UdS test sets. UD-token surprisal and BLEU scores improve after fine-tuning (as expected), but cross-lingual surprisal from MT models is worse.}

\end{table*}

\begin{figure*}[!t]
\centering
\section*{Appendix D. Extended Case Study Illustrations}
\vspace{0.5em}

\includegraphics[width=\linewidth]{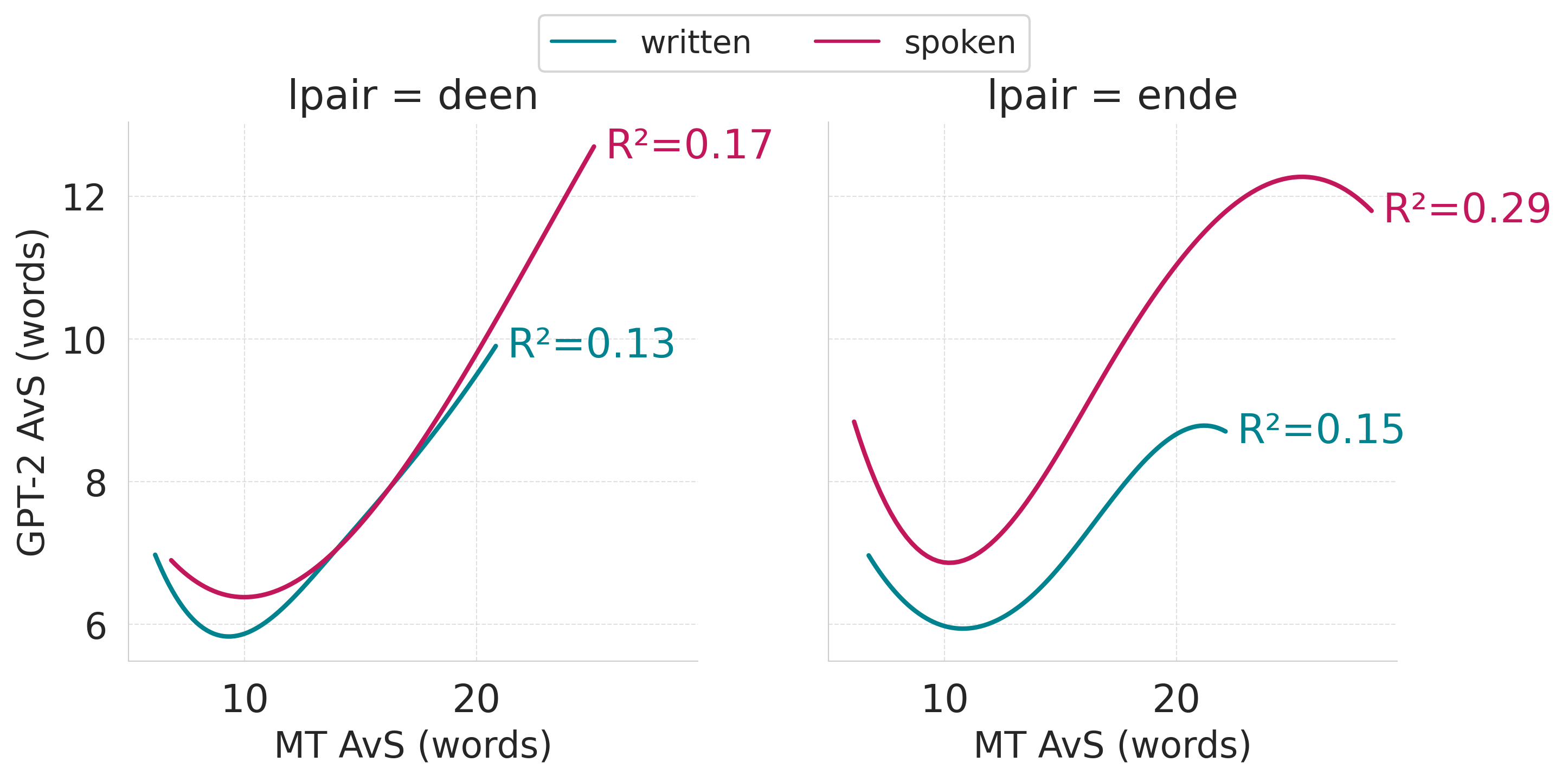}

\caption{\label{fig:gam}
Fitted linear GAM regression lines and pseudo $R^2$ statistics.
Setup: n\_splines=5, gridsearch run on word surprisals from base models for test sets in EuroParl-EPIC-UdS.}
\end{figure*}

\end{document}